\documentclass[10pt,twocolumn,letterpaper]{article}

\usepackage{cvpr}
\usepackage{times}
\usepackage{epsfig}
\usepackage{graphicx}
\usepackage{amsmath}
\usepackage{amssymb}

\usepackage{booktabs}
\usepackage{multirow}
\usepackage{float}
\usepackage{amsmath}
\usepackage[skip=0pt]{caption}
\usepackage{rotating}
\usepackage{subcaption}

\usepackage[pagebackref=true,breaklinks=true,letterpaper=true,colorlinks,bookmarks=false]{hyperref}

\cvprfinalcopy 

\def\cvprPaperID{3753} 

\ifcvprfinal\fi
\begin{document}

\title{Dynamics Transfer GAN:  Generating Video by Transferring Arbitrary Temporal Dynamics from a Source Video to a Single Target Image}

\author{Wissam J. Baddar, Geonmo Gu, Sangmin Lee and Yong Man Ro
\\
Image and Video Systems Lab., School of Electrical Engineering, KAIST, South Korea\\
{\tt\small \{wisam.baddar,geonm,sangmin.lee,ymro\}@kaist.ac.kr}}

\maketitle

\maketitle
\begin{abstract}
In this paper, we propose Dynamics Transfer GAN; a new method for generating video sequences based on generative adversarial learning. The spatial constructs of a generated video sequence are acquired from the target image. The dynamics of the generated video sequence are imported from a source video sequence, with arbitrary motion, and imposed onto the target image. To preserve the spatial construct of the target image, the appearance of the source video sequence is suppressed and only the dynamics are obtained before being imposed onto the target image. That is achieved using the proposed appearance suppressed dynamics feature. Moreover, the spatial and temporal consistencies of the generated video sequence are verified via two discriminator networks. One discriminator validates the fidelity of the generated frames appearance, while the other validates the dynamic consistency of the generated video sequence. Experiments have been conducted to verify the quality of the video sequences generated by the proposed method. The results verified that Dynamics Transfer GAN successfully transferred arbitrary dynamics of the source video sequence onto a target image when generating the output video sequence. The experimental results also showed that Dynamics Transfer GAN maintained the spatial constructs (appearance) of the target image while generating spatially and temporally consistent video sequences.
\end{abstract}
\vspace{-0.5cm}
\section{Introduction}
\begin{figure}[!ht]
\begin{center}
\includegraphics[width=1.0\linewidth] {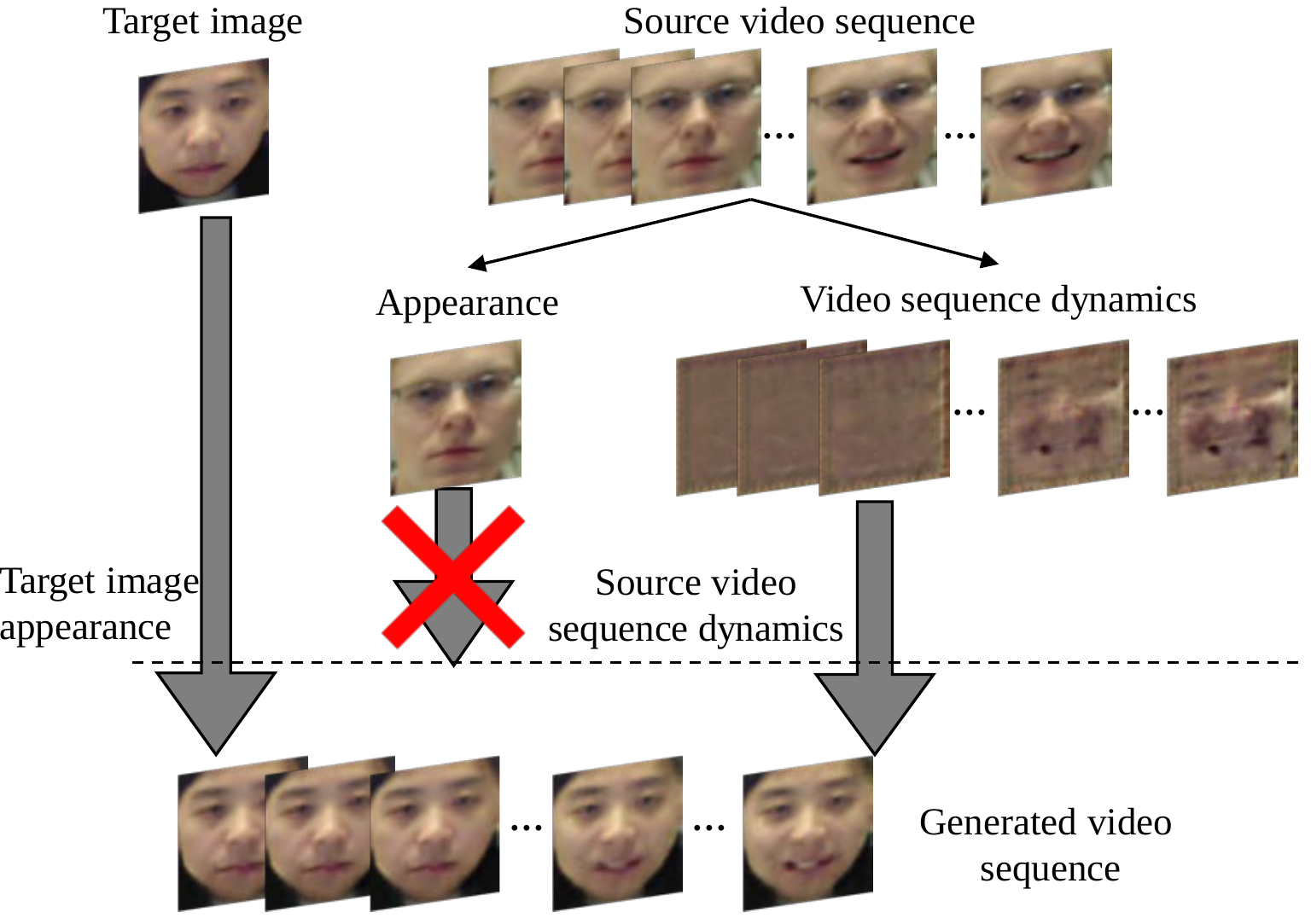}
\end{center}
   \caption{Proposed Dynamics Transfer GAN. Given a single target image and a video sequence with arbitrary temporal dynamics, the proposed Dynamics Transfer GAN generates a synthesized video sequence with the dynamics of the source video onto the appearance of the target image.}
\label{fig:1}
\vspace{-0.5cm}
\end{figure}

The recent advances in generative models have influenced researches to investigate image synthesis. Generative models, particularly generative adversarial networks (GANs), have been utilized to generate images from random distributions \cite{1,2}, or synthesize images by non-linearly transforming a priming image to the synthesized image \cite{3,4,5}, or even synthesizing images from a source image domain to a different domain \cite{6,7,8}.

The progress towards better image generation has been witnessing an interesting surge  \cite{7,8,9,10,11,12}. Extending the capacities of generative models to generate video sequences is the natural and inevitable progression. However, extending generative models to generate meaningful video sequences is a challenging task. Generating meaningful video sequences requires the generative model to understand the spatial constructs of the scene as well as the temporal dynamics that drive the scene motions. In addition, the generative model should be able to reconstruct temporal variations with variable sequence lengths. In many cases, the dynamic motion could be non-rigid or cause shape transformation of the underlying spatial construct. As such, many aforementioned aspects could hamper the effectiveness of generative models to generate videos.

Due to the challenges mentioned above, some research efforts have tried to simplify the problem by limiting the generation to predict a few future frames of a given video sequence \cite{13,14,15,16,17,18,19}. In these works, combinations of 3D convolutions and recurrent neural networks (RNNs) and convolutional long-short-term-memory (LSTM) were investigated to predict future frames. While many of these methods have shown impressive and promising results, predicting a few future frames is considered as a conditional image generation problem, which is different from the video generation \cite{20}.

The authors in \cite{21} proposed an extension to GANs that generates videos with scene dynamics. The generator was composed of two streams to model the scene as a combination of a foreground and a background. 3D convolutions were utilized to perform spatio-temporal discriminators that criticize the generated sequences. A similar two-stream generator one spatio-temporal discriminator approach was proposed in \cite{22}. Both \cite{21,22} could not model variable length video sequences, and could not generate long sequences. The authors of \cite{20} separated the sampling procedure for input distribution into samples from a content subspace and a motion subspace to generate variable length sequences. The works in \cite{20,21,22} showed that GANs could be extended to generate videos. However, the spatio-temporal discriminator was performed using 3D convolutions of fixed size, which meant that the spatio-temporal consistency of the generated videos could be limitedly verified at a fixed small sequence size. Moreover, the motion was coupled with the spatial construct in the spatio-temporal encoding process, which could limit the ability to generate dynamics at the desired spatial appearance.

In this paper, we propose a new video generation method named as Dynamics Transfer GAN. The proposed video generation is primed with a target image. The video sequence is generated by transferring the dynamics of arbitrary motion from a source video sequence onto the target image. The main contributions of the proposed method are summarized as follows:  

\begin{enumerate}
\item We propose a new video sequence generation method from a single target image. The target image holds the spatial construct of the generated video sequence. The dynamics of the generated video are imported from an arbitrary motion of a source video sequence. The proposed method maintains the spatial appearance of the target image while importing the dynamics from a source video. To that end, we propose new appearance suppressed dynamics feature, which suppresses the spatial appearance of the source video while maintaining the temporal dynamics of source sequence. The proposed dynamics feature is devised so that the effect of spatial appearance is suppressed in the spatio-temporal encoding with a RNN. Thus, in video sequence generation, the source video dynamics are imported and imposed onto the target image.

\item The proposed Dynamics Transfer GAN is designed with the goal of generating variable length video sequences that extend in time (i.e., no limitation on the sequence length). To that end, we design a generator network with two discriminator networks. One discriminator investigates the fidelity of the generated frames (spatial discriminator). The other discriminator investigates the integrity of the generated sequence as a whole sequence (dynamics discriminator). In longer sequences, it could be expected that the dynamics discriminator focuses on the tailing parts of a video. To continuously maintain the spatial and dynamic fidelity of the generated video, additional objective terms were added to the training of the generator network. As a result, the proposed method generates videos with realistic spatial structure of the target image and temporal dynamics that mimic the source video sequence. 

\item We provide a visualization of the imported dynamics from the source video. The dynamics visualization helps in understanding how the generative parts of the network perceive the input video sequence dynamics. Moreover, the visualization demonstrates that the proposed dynamics feature suppresses the source video appearance and only encodes the dynamics of the source video.
\end{enumerate}

\section{Related Work}

\subsection{Generative Adversarial Networks}
Generative adversarial networks (GANs) have been proposed in \cite{1} as a 2-player zero-sum game problem consisting of two networks: a generator network and a discriminator network. The generator network $(G: \mathbb{R}^K \rightarrow \mathbb{R}^M)$ tries to generate a sample $(\widehat{\textbf{x}}\in \mathbb{R}^M)$ which mimics a sample in a given dataset $(\textbf{x} \in \mathbb{R}^M)$. As an input, the generator network receives a latent variable $(\textbf{z} \in \mathbb{R}^K)$, which is randomly sampled from a given distribution $p_{\textbf{z}}$. Different distributions have been proposed to model the distribution of the latent variable $p_{\textbf{z}}$, such as a Gaussian model \cite{1}, a mixture of Gaussians model \cite{2} or even in the form of dropout to an input image \cite{7}. On the other hand, the goal of the discriminator $(D: \mathbb{R}^M\rightarrow [0,1])$ is to investigate the fidelity of the sample, and try to distinguish whether the given sample is real (ground truth sample) or fake (generated sample).

Training GANs can be achieved by simultaneously training the generator $(G)$ and discriminator $(D)$ networks with a non-cooperative game; i.e., the discriminator wins when it correctly distinguishes fake samples from real samples, while the generator wins when it generates samples that can fool the discriminator. Explicitly, the training of $G$ and $D$ is achieved by solving the minimax game with the value function:
\vspace{-0.2cm}
\begin{align}
&\underset{G}{\min}\,\underset{D}{\max} \, \mathcal{L}(D,G) = \mathbb{E}_{\textbf{x} \sim p_{\textbf{x}}}[\log{D(\textbf{x}))}] \nonumber\\ 
&+\mathbb{E}_{{\textbf{z}} \sim p_{\textbf{z}}}[\log{(1-D(\textbf{z}))}].
\end{align}
\label{eq:1}

\subsection{Video Generation with GANs}
Extending GAN to video generation is an instinctive progression from GAN for image generation. However, a few methods have tried generating complete video sequences \cite{20,21,22,23}. The authors of both \cite{21,22} have proposed a two-stream generator and one spatio-temporal discriminator approach. The authors in \cite{21} assumed that a video sequence was constructed of a foreground and background and they separated generators accordingly. In \cite{22}, the video sequence was modeled by an image stream and a temporal stream. A limitation of these works is that the generator could limitedly generate fixed short-length video sequences.

\begin{figure} [!ht]
    \centering
    
    \begin{subfigure}{1.0\linewidth}
    	\centering
        	\setlength{\abovecaptionskip}{1.0 pt}
            \includegraphics[width=1.0\linewidth,height=3cm,keepaspectratio]{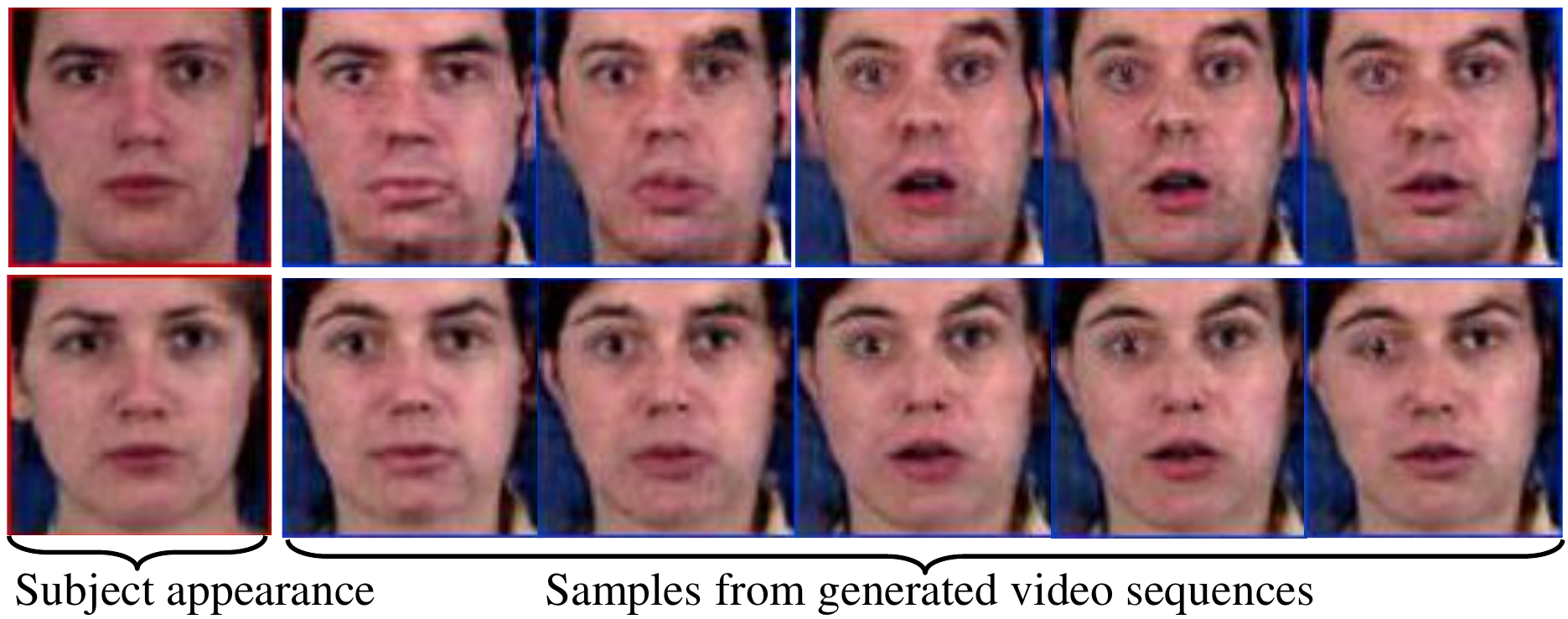}
            \caption{Examples of generated video frames using \cite{20}.}
			\label{fig2:a}
    \end{subfigure}

	\begin{subfigure}{1.0\linewidth}
		\centering
        	\setlength{\abovecaptionskip}{1.0 pt}
			\includegraphics[width=1.0\linewidth,height=3cm,keepaspectratio]{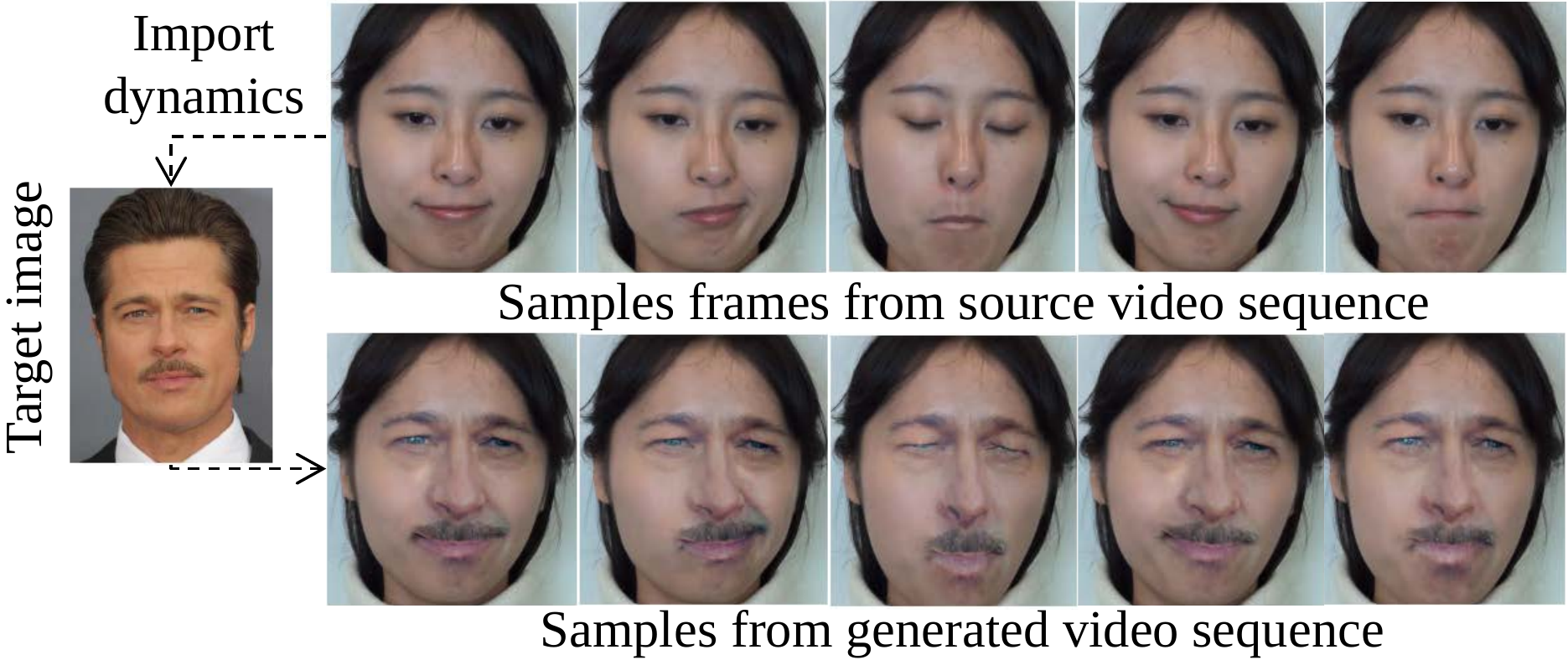}
			\caption{Examples of generated video frames using \cite{23}.}
			\label{fig2:b}
     \end{subfigure}

\caption{Examples of severe spatial-construct artifacts in previously proposed video generation methods. }
\label{fig:2}
\vspace{-0.5cm}
\end{figure}

The authors in \cite{20} proposed MoCoGAN, a GAN for generating video sequences without a priming image. MocoGAN divided input distribution into content subspace and motion subspace. The sampling from the content subspace was accomplished by sampling from a Gaussian distribution. The sampling from the motion subspace was performed using an RNN. As such a content discriminator and motion discriminator were developed. MoCoGAN could generate sequences with variable lengths. However, the motion discriminator was limited to handle a fixed number of frames. This meant that the motion consistency of the generated videos was limitedly verified on a limited number of frames. Figure~\ref{fig2:a} shows examples of frames generated via MoCoGAN \cite{20}, in which the content of the generated video sequences was set to different subject appearances, while the motion was fixed to the same expression. As shown in the figure, the appearance of the generated video sequences is quite similar although the content (spatial construct) was different. In fact, in both cases, the subject identity of generated frames was fairly changed.

The authors in \cite{23} proposed importing the dynamics from a source video sequence to a target image. However, this method resulted in severe disruption in the appearance of the target image. Figure~\ref{fig2:b} shows examples of frames taken from generated video sequences using \cite{23}. It is clear from the figure that the method in \cite{23} could capture the dynamics of the source video sequence. However, the spatial construct of the target image is severely damaged. The generated sequences follow the facial structure of the source sequence and append textural features of the target image onto it. In our proposed method, we intend to transfer the dynamics from the source video to the generated video while maintaining the appearance of the target image.

\section{The Proposed Dynamics Transfer GAN}

Given an image $\textbf{x}$ and a source video sequence $\textbf{Y}=[\textbf{y}_0,\textbf{y}_1,...,\textbf{y}_t,...,\textbf{y}_\textup{T}]$ with a frame $\textbf{y}_i$ and sequence length $\textup{T}$, the objective of the proposed method is to import the temporal dynamics from a source video and impose the dynamics on the input target image $\textbf{x}$. As a result, the generator should generate a video sequence $\widehat{\textbf{\textup{Y}}}=[\widehat{\textbf{y}}_0,\widehat{\textbf{y}}_1,...\widehat{\textbf{y}}_t,...,\widehat{\textbf{y}}_\textup{T}]$ of length $\textup{T}$ and generated frame $\widehat{\textbf{y}}_i$. The generated video sequence is supposed to possess the appearance of target image $\textbf{x}$ and the dynamic of source sequence $\textbf{Y}$. In the following subsections, we detail the proposed Dynamics Transfer GAN. First, we detail the proposed method for obtaining the input of Dynamics Transfer GAN. Then, the proposed GAN network structure and the training procedure with the proposed objective terms are explained.

\subsection{Input of Dynamics Transfer GAN}
The input of a video generative model could be represented as a vector of $\textup{T}$ samples in the latent space denoted as $(\textbf{\textup{Z}}=[\textbf{z}_0,\textbf{z}_1,...,\textbf{z}_t,...,\textbf{z}_\textup{T}])$ \cite{20}. Each sample $\textbf{z}_t$  in $\textbf{\textup{Z}}$ represents a frame at time $t$. By traversing the samples of the vector $\textbf{\textup{Z}}$, we can explore the temporal path in which the video sequence traverses through. At any point in time $t$, $\textbf{z}_t$ is decomposed into a spatial representation $(\textbf{z}_{t}^{(s)})$ and a temporal dynamics representation $(\textbf{z}_{t}^{(d)})$. 

In this paper, we fix the spatial representation as the target image$(\textbf{z}_{t}^{(s)} = \textbf{x})$, such that the spatial appearance follows the target image appearance. Note that when generating the spatial representation for a sample $\textbf{z}_{t}^{(s)}$, instead of adding random noise to the target image $\textbf{x}$, the noise is provided in the form of dropout applied on several layers of the generator as described in \cite{7}. The dynamic representation, on the other hand, is attained from an appearance suppressed dynamics feature obtained using a pre-trained RNN \cite{24}). The role of the RNN is to obtain a spatio-temporal representation of each frame in the source sequence. This allows us to represent dynamics at each frame as a sample point in the generator input space. Further, the RNN facilitates generating arbitrary length videos. In this paper, when imposing the dynamics of the source video sequence to the target image, we are interested in imposing the dynamics while maintaining the target image appearance. We devise appearance suppressed dynamics feature to eliminate the effect of spatial encoding of the source video in the pre-trained RNN.

\begin{figure}
\begin{center}
\includegraphics[width=0.8\linewidth] {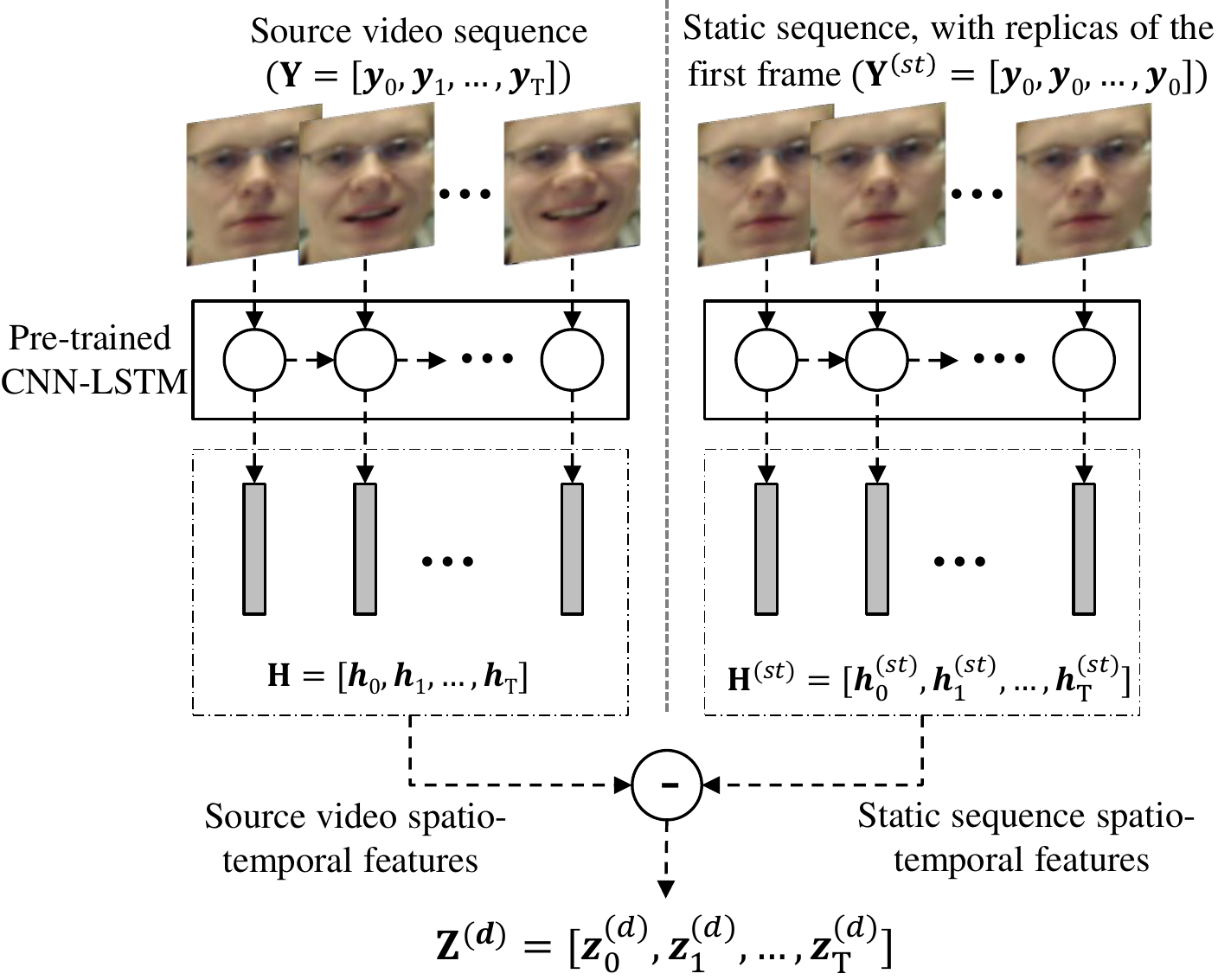}
\end{center}
   \caption{Appearance suppressed dynamics feature encoder for the input of Dynamics Transfer GAN.}

\label{fig:3}
\vspace{-0.6cm}
\end{figure}

Figure~\ref{fig:3} details the proposed method to obtain the appearance suppressed dynamics feature $\textup{\textbf{Z}}^{(d)}$. As shown in the figure, from the source video sequence $\textbf{\textup{Y}}$, a static sequence $\textbf{\textup{Y}}^{(st)}=[\textbf{y}_0,\textbf{y}_0,...,\textbf{y}_\textup{0}]$ is generated by replicating the first frame of the source video sequence $\textup{T}$ times. Both the source video sequence $\textbf{\textup{Y}}$ and the static sequence $\textbf{\textup{Y}}^{(st)}$ are fed into the pre-trained RNN to generate the source video latent spatio-temporal features $\textup{\textbf{H}}$ and the static sequence latent spatio-temporal features $\textup{\textbf{H}}^{(st)}$, respectively. Since $\textup{\textbf{Y}}^{(st)}$ is $\textup{T}$ replicas of the same image, the RNN only encodes the spatial appearance in $\textup{\textbf{H}}^{(st)}$ rather than temporal features. Thus, by subtracting $\textup{\textbf{H}}^{(st)}$ from $\textup{\textbf{H}}$, the spatial appearance of the source video is suppressed and the dynamics of the source video are disentangled. Namely, the appearance suppressed dynamics feature can be obtained by $\textup{\textbf{Z}}^{(d)}=\textup{\textbf{H}}-\textup{\textbf{H}}^{(st)}$(see to the visualization of the appearance suppressed dynamics feature in Figure~\ref{fig:7}).
\begin{figure*}[!ht]
\begin{center}
\includegraphics[width=0.78\linewidth] {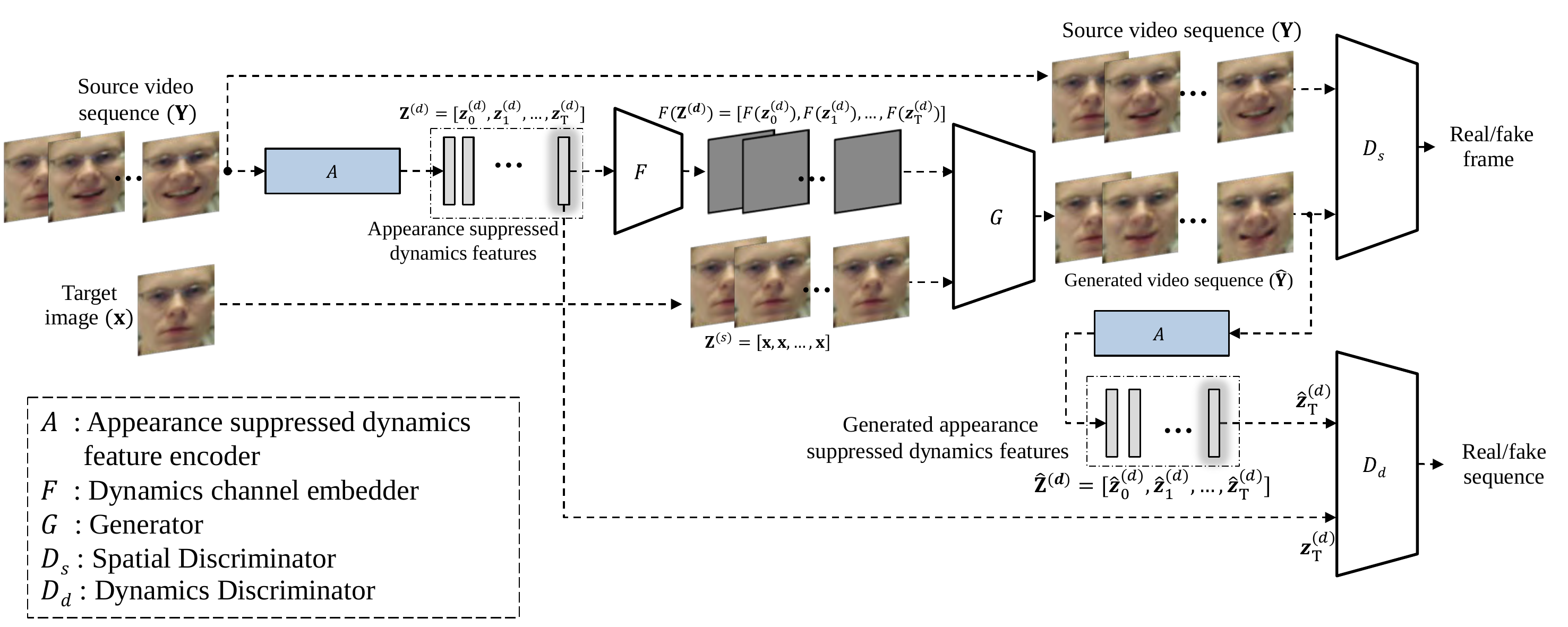}
\end{center}
   \caption{Overview of the proposed Dynamics Transfer GAN}
   \vspace{-0.5cm}
\label{fig:4}
\end{figure*}

\subsection{The Proposed Dynamics Transfer GAN}
Figure~\ref{fig:4} shows an overview of the proposed Dynamics Transfer GAN. The proposed Dynamics Transfer GAN is constructed of five main components: appearance suppressed dynamics feature encoder $(A)$, dynamics channel embedder $(F)$, generator network $(G)$, spatial discriminator network $(D_s)$ and a dynamics discriminator network $(D_d)$. 

The appearance suppressed dynamics feature encoder (mentioned in section 3.1) encodes the temporal dynamics feature of a source video. In this paper, to encode the dynamics effectively, the RNN model parameters employed from \cite{24} are frozen during the training stage of the proposed Dynamics Transfer GAN.

The proposed Dynamics Transfer GAN generates a video sequence by synthesizing a sequence of frames. Hence both the spatial representation $(\textbf{z}_{t}^{(s)})$ and the dynamic representation $(\textbf{z}_{t}^{(d)})$ described in section 3.1 are fed into the generator at each frame. To combine the spatial representation $(\textbf{z}_{t}^{(s)})$ with the dynamic representation $(\textbf{z}_{t}^{(d)})$ at a time $t$, we embed the dynamics feature representation $(\textbf{z}_{t}^{(d)})$ into a feature channel by using the dynamics feature embedder network $(F)$. The embedded feature channel is concatenated with the target image $(\textbf{z}_{t}^{(s)}=\textbf{x})$ and fed to the generator. Note that we also use dropout on the dynamics feature embedder network to generate noise to the dynamics representation. 

To maintain the target image appearance in the synthesized video, the generator network structure should be able to preserve the target image appearance. Many previous works in constructing a fine detailed image have utilized some form of encoder-decoder networks \cite{25,26,27}. In this paper, we employ a U-net network structure for the generator network which could preserve details in image generator network \cite{7,28}. The generator input is the concatenation of the spatial representation $(\textup{\textbf{Z}}^{(s)})$ and the embedded dynamics feature representation channel $F(\textup{\textbf{Z}}^{(d)})$. The generator allows for a variable length video sequences $(\textup{\textbf{Y}})$ to be fed into it, so that it could generate a variable length sequence $(\widehat{\textup{\textbf{Y}}})$. 

To criticize and discriminate the generated images, two discriminator networks are devised: spatial discriminator network $(D_s)$ and dynamics discriminator network $(D_d)$. The goal of the spatial discriminator is to check the fidelity of each generated frame, and try to distinguish real frames from generated frames. The spatial discriminator network structure is fairly straightforward. It is constructed of a stack of convolutional networks and an output layer for discriminating whether each frame is a real frame or a generated (fake) frame.

The purpose behind the dynamics discriminator is to distinguish if the dynamics of the generated sequence represent realistic dynamics or fake dynamics. The appearance suppressed dynamic feature for the generated sequence $(\widehat{\textup{\textbf{Y}}})$ is obtained by the appearance suppressed dynamics feature encoder $(A)$.  Similar to details described in Figure~\ref{fig:3}, to suppress the effect of appearance in the generated sequence, the RNN is utilized to obtain the generated static-latent features $(\widehat{\textup{\textbf{H}}}^{(st)})$ from a static sequence $\widehat{\textbf{\textup{Y}}}^{(st)}=[\widehat{\textbf{y}}_0,\widehat{\textbf{y}}_0,..., \widehat{\textbf{y}}_0]$. Accordingly, the generated appearance suppressed dynamics feature can be obtained by $(\widehat{\textup{\textbf{Z}}}^{(d)}= \widehat{\textup{\textbf{H}}}-\widehat{\textup{\textbf{H}}}^{(st)})$. At the dynamics discriminator, the realistic dynamics of a sequence is donated by $(\textup{\textbf{Z}}^{(d)})$ and fake (generated) sequence dynamics is denoted by $(\widehat{\textup{\textbf{Z}}}^{(d)})$.

It should be noted that the dynamics discriminator and the spatial discriminator deal with the generated video differently. The spatial discriminator views generated video as a sequence of frames so that each frame represents a sample in the input space of spatial discriminator. The dynamics discriminator views the generated video sequence as a sample in the input space of dynamics discriminator. Note that a video sequence could have a variable length. To allow the dynamics discriminator to deal with the whole sequence as a sample point regardless of the sequence length, the input size of the dynamics discriminator should not be affected by the length of the sequence. Therefore, we only utilize the appearance suppressed feature $(\widehat{\textbf{z}}_\textup{T}^{(d)})$ at time $\textup{T}$. Due to the RNN properties, $(\widehat{\textbf{z}}_\textup{T}^{(d)})$ represents the dynamics of full sequence, i.e., from the beginning until time $\textup{T}$.

\subsection{Training the Proposed Dynamics Transfer GAN}
The training of the Dynamics Transfer GAN is a 2-player zero-sum game problem. Specifically, in this paper, the generative part of the network is a group of networks, i.e., the dynamics channel embedder $(F)$ and the generator network $(G)$. The discriminative part of Dynamics Transfer GAN is both the spatial discriminator $(D_s)$ and the dynamics discriminator $(D_d)$ networks. Note that the RNN parameters in the appearance suppressed dynamics feature encoder $(A)$ are frozen. Thus, they are not included in the gradient update process. Explicitly, the training of $F, G, D_s$, and $D_d$ is achieved by solving the minimax problem with the value function:
\vspace{-0.2cm}
\begin{align}
&\underset{F,G}{\min}\,\underset{D_s,D_d}{\max} \, \mathcal{L}(F,G,D_s,D_d) = 
\mathbb{E}_{\textbf{y} \sim p_{\textbf{y}} }[\log{(D_s(\textup{\textbf{Y})}}] \nonumber\\
&+ \mathbb{E}_{\textbf{z}^{(d)} \sim p_{\textbf{z}}^{(d)} }[\log{(D_d(\textbf{z}}_{\textup{T}}^{(d)}))]  \nonumber\\
&+ \mathbb{E}_{\textbf{Z}^{(d)} \sim p_{\textbf{Z}}^{(d)},\textbf{Z}^{(s)} \sim p_{\textup{\textbf{Z}}}^{(s)} }[\log{(1-D_s(G(\textup{\textbf{Z}}}^{(s)},F(\textup{\textbf{Z}}^{(d)}))))]  \nonumber\\ 
&+ \mathbb{E}_{\widehat{\textbf{z}}^{(d)} \sim p_{\widehat{\textbf{z}}}^{(d)} }[\log{(1-D_d(\widehat{\textbf{z}}}_{\textup{T}}^{(d)}))].
\end{align}
\label{eq:2}
\vspace{-0.5cm}
       
Note that the back propagation can be performed independently on the discriminator networks $(D_d, D_s)$. The generative networks $(F, G)$ are updated after both discriminators $(D_d, D_s)$. Therefore, in practice, the training of the discriminators $(D_d, D_s)$ and that of the generative networks $(F, G)$ are performed alternatively. In the first step, discriminators $(D_d, D_s)$ are trained by maximizing the loss terms:
\vspace{-0.3cm}
\begin{align}
&\mathcal{L}_{D_s}(F,G,D_s) = \mathbb{E}_{\textbf{y} \sim p_{\textbf{y}} }[\log{(D_s(\textup{\textbf{Y})}}] \nonumber\\
&+ \mathbb{E}_{\textbf{Z}^{(d)} \sim {p_{\textbf{Z}^{(d)}}}, \textbf{Z}^{(s)} \sim {p_{\textbf{Z}^{(s)}}} }[\log{(1-D_s(G(\textup{\textbf{Z}}}^{(s)},F(\textup{\textbf{Z}}^{(d)}))))] , \nonumber\\ 
&\mathcal{L}_{D_d}(F,G,D_d) = \mathbb{E}_{\textbf{z}^{(d)} \sim p_{\textbf{z}^{(d)}} }[\log{(D_d(\textbf{z}}_{\textup{T}}^{(d)}))]  \nonumber\\
&+ \mathbb{E}_{\widehat{\textbf{z}}^{(d)} \sim {p_{\widehat{\textbf{z}}^{(d)}}} }[\log{(1-D_d(\widehat{\textbf{z}}}_{\textup{T}}^{(d)}))].
\end{align}
\label{eq:3}
In the second step, the generative parts of the network are trained by minimizing the adversarial loss: 
\vspace{-0.2cm}
\begin{align}
&\mathcal{L}_{G^{(A)}}(F,G,D_s,D_d) = \mathbb{E}_{\widehat{\textbf{z}}^{(d)} \sim p_{\widehat{\textbf{z}}^{(d)}} }[-\log{(D_d(\widehat{\textbf{z}}}^{(d)}))] \nonumber\\
& + \mathbb{E}_{\textbf{Z}^{(d)} \sim p_{\textbf{Z}^{(d)}},\textbf{Z}^{(s)} \sim p_{\textbf{Z}^{(s)}} }[-\log{(D_s(G(\textup{\textbf{Z}}}^{(s)},F(\textup{\textbf{Z}}^{(d)}))))].
\end{align}
\label{eq:4}
\vspace{-0.5cm}

Adding a reconstruction term to the generative networks could improve the quality of the generated images \cite{7}, We employ L1 frame reconstruction loss to improve the spatial reconstruction at each frame as follows:
\vspace{-0.2cm}
\begin{align}
\mathcal{L}_{G^{(s)}}(F,G) = \mathbb{E}_{\textbf{y} \sim p_{\textbf{y}}} \Big[\left \| \textbf{Y}- G(\textbf{Z}^{(s)}, F(\textbf{Z}^{(d)})) \right \|_{1}\Big].  
\end{align}
\label{eq:5}
\vspace{-0.5cm}	

Finally, to maintain dynamic consistency even when the sequence is lengthy, we propose a reconstruction term on the appearance suppressed dynamics feature. Unlike the loss from the dynamics discriminator (which is calculated at the end of the sequence), this term makes sure that the generated sequence dynamics are maintained at each frame. 
\vspace{-0.2cm}
\begin{align}
\mathcal{L}_{G^{(d)}}(F,G) = \mathbb{E}_{\textup{\textbf{Z}}^{(d)} \sim p_{\textup{\textbf{Z}}^{(d)}}} \Big[\left \| \textup{\textbf{Z}}^{(d)}- \widehat{\textup{\textbf{Z}}}^{(d)} \right \|_{1}\Big].  
\end{align}
\label{eq:6}
\vspace{-0.5cm}

By plugging in eqs. (5) and (6) in the second step of the training described in eq.(4), the generator training can be obtained by minimizing the loss:
\vspace{-0.1cm}
\begin{align}
&\mathcal{L}_{G}(F,G,D_s,D_d)  = \lambda_{G^{(A)}} \mathcal{L}_{G^{(A)}}(F,G,D_s,D_d) \nonumber\\
&+ \lambda_{G^{(s)}} \mathcal{L}_{G^{(s)}}(F,G)
+ \lambda_{G^{(d)}} \mathcal{L}_{G^{(d)}}(F,G),
\end{align}
\label{eq:7}
where $\lambda_{G^{(A)}}$, $\lambda_{G^{(s)}}$, and $\lambda_{G^{(d)}}$  are hyper-parameters to control the generative loss term. $\lambda_{G^{(A)}}$ is the weight controlling the adversarial loss of the Dynamics Transfer GAN described in eq. (4), $\lambda_{G^{(s)}}$ is the weight controlling the frame reconstruction loss detailed in eq. (5), and $\lambda_{G^{(d)}}$ is the weight controlling the loss that maintains the dynamic consistency at each frame described in eq. (6).


\section{Experiments}

\subsection{Experiment Setup}
To verify the effectiveness of the proposed Dynamics Transfer GAN, experiments were conducted on the Oulu-CASIA dataset \cite{34}. In the dataset, sequences of the six basic expressions (i.e., angry, disgust, fear, happy, sad, and surprise) were collected from 80 subjects under three illumination conditions. For the experiments, video sequences collected with a visible light camera under normal illumination conditions were used. A total of 480 video sequences were collected (6 video sequences per subject, 1 video sequence per expression). For each subject, the basic expression sequence was captured from a neutral face until the expression apex. In the experiments, the face region was detected and facial landmark detection was performed \cite{35} on each frame. The face region was then automatically cropped and aligned based on the eye landmarks \cite{36}. The training and implementation of the proposed Dynamics Transfer GAN have been conducted using TensorFlow \cite{29}. 

\begin{figure*}[!ht]
\begin{center}
\includegraphics[width=1\linewidth] {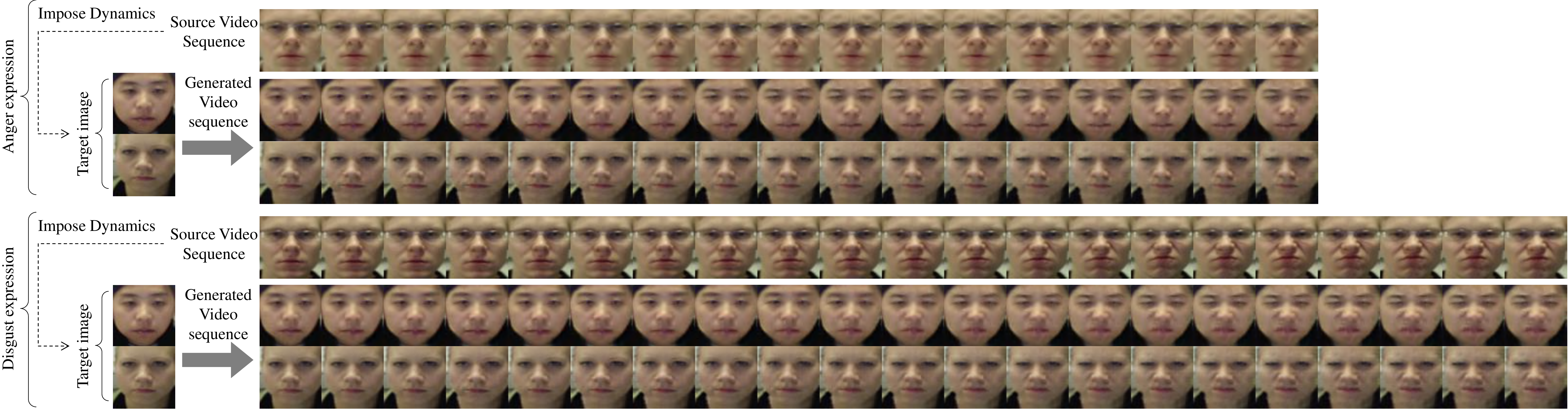}
\end{center}
   \caption{Example of sequences of arbitrary length generated by the proposed Dynamics Transfer GAN.}
   \vspace{-0.5cm}
\label{fig:5}
\end{figure*}

\subsection{Experiment 1: Visual Qualitative Results of Videos Generated with the Proposed Dynamics Transfer GAN}
In experiment 1, we investigated the quality of the generated video sequences by using the proposed Dynamics Transfer GAN. Figure~\ref{fig:5} shows examples of generated sequences to qualitatively assess the generated video sequence (more examples of generated video sequences can be found in the appendix, section 6.1). The figure shows the source video sequence, the target image and the corresponding generated video sequences. As seen in the figure, the generated videos are of arbitrary length, and the length of the sequence has no effect on the quality of the generated image. This is attributed to the dynamics discriminator, and the generator dynamics objective term. From the figure, it can also be seen that multiple videos could be generated from the same source video sequence. This was achieved by imposing the dynamics of the source video sequence onto different target images. The figure also presents an example of the same appearance with different dynamics (e.g., different expression sequence imposed on the same target image). This was accomplished by fixing the target image, while changing the source video sequence. These results show that the proposed method effectively transfers arbitrary dynamics of source video sequences to the generated video sequences. In addition, the appearance and the identity of the target image were preserved in the generated video sequences.

\subsection{Experiment 2: Subjective Assessment of Videos Generated with the Proposed Dynamics Transfer GAN}
It is known that quantitatively evaluating generative models, especially on generating visual outputs, is challenging \cite{20}. It is also shown that all the popular metrics are subject to flaws \cite{37}. Therefore, in experiment 2, we performed a subjective experiment in order to quantitatively evaluate the generated video sequences \cite{20}. To that end, 10 participants took part of the experiment to evaluate the quality of the generated videos. The participants viewed a total of 240 generated video sequences. The video sequences were generated by transferring the dynamics from 24 source video sequences into 10 target images. The source sequences represented the 6 basic facial expressions (i.e., anger, disgust, fear, happiness, sadness and surprise) of 4 subjects. The target images were neutral expression images of 10 subjects. When evaluating the generated video sequences, the participants were guided to watch the generated video sequences along with the corresponding source video sequences and target images. After viewing each sequence, the subjects were asked to rate if the generated video is realistic or not. Table~\ref{Table:1} shows the percentage of the video sequences that were rated as realistic. As can be seen from the results, more that 78.33\% of the videos were rated as realistic. This percentage reflects that the proposed Dynamics Transfer GAN has generated video sequences with a reasonable quality.

In evaluating the quality of the generated videos, it is important to make sure that (1) the generated videos preserved the appearance of the target image, and that (2) the dynamic transitions in the video sequence should be smooth and consistent. To that end, each subject in the subjective assessment experiment was asked to rate the spatial consistency of the generated video sequence (i.e., the spatial appearance of the target image was preserved, and the spatial constructs in the generated frames was intact). Added to that, the subjects were asked to rate if the generated videos were temporally consistent (i.e., the dynamic transition between the frames was smooth). The results of the subjective evaluation are shown in Table~\ref{Table:1}. The results show that the proposed Dynamics Transfer GAN could generate temporally consistent sequences while preserving the appearance of the target image. 

\begin{table}[!ht]
  \centering
  \caption{Subjective quality assessment results for the generated video sequence}
  \label{Table:1}
  \resizebox{\columnwidth}{!}{
    \begin{tabular}{p{8.33cm}rrr}
    \toprule
    \toprule
    \multicolumn{1}{p{4.73 cm}}{\textbf{Question}}  & \multicolumn{1}{p{1.4 cm}}{\textbf{Yes(\%)}} & \multicolumn{1}{p{1.4cm}}{\textbf{No(\%)}} \\
   
    \midrule
    \multicolumn{1}{p{4.73cm}}{Is this video realistic } & \multicolumn{1}{c}{78.33\%} & \multicolumn{1}{c}{21.67\%} \\
    \multicolumn{1}{p{4.73cm}}{Is the video temporally consistent} & \multicolumn{1}{c}{82.92\%} & \multicolumn{1}{c}{17.08\%} \\
    \multicolumn{1}{p{4.73cm}}{Is the video spatially consistent} & \multicolumn{1}{c}{79.17\%} & \multicolumn{1}{c}{20.83\%} \\
    \midrule
    \midrule

    \end{tabular}%
    }
\end{table}%
\vspace{-0.5cm}

\begin{figure} [!ht]
    \centering
    
    \begin{subfigure}{1.0\linewidth}
    	\centering
        	\setlength{\abovecaptionskip}{1.0 pt}
            \includegraphics[width=1.0\linewidth,height=1.1cm,keepaspectratio]{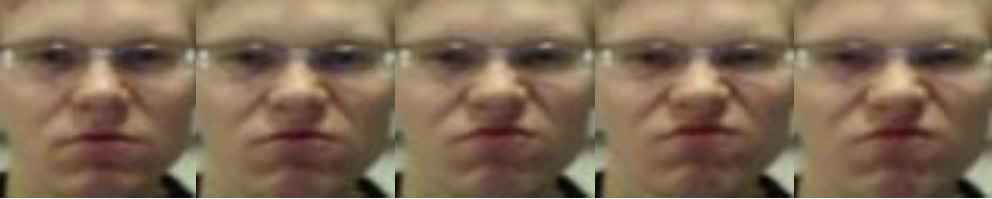}
            \caption{Samples from the source video sequence.}
			\label{fig6:a}
    \end{subfigure}
          
	\begin{subfigure}{1.0\linewidth}
		\centering
        	\setlength{\abovecaptionskip}{1.0 pt}
			\includegraphics[width=1.0\linewidth,height=2.2cm,keepaspectratio]{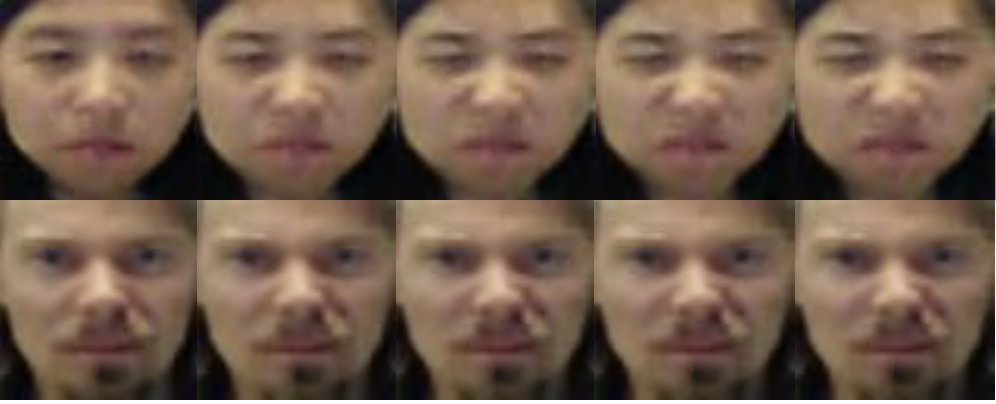}
			\caption{Samples generated with Dynamics Transfer GAN using the proposed appearance suppressed Dynamics feature.}
			\label{fig6:b}
     \end{subfigure}
     
	\begin{subfigure}{1.0\linewidth}
		\centering
        	\setlength{\abovecaptionskip}{1.0 pt}
			\includegraphics[width=1.0\linewidth,height=2.2cm,keepaspectratio]{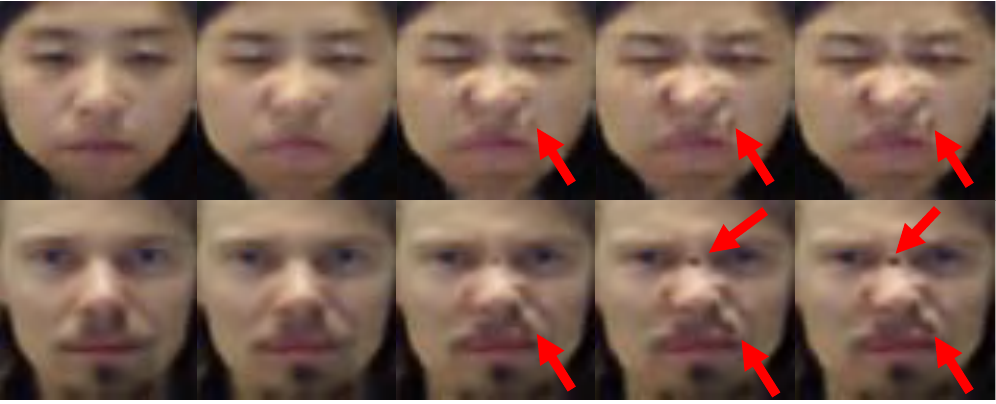}
			\caption{Samples generated with Dynamics Transfer GAN using the CNN-LSTM features.}
			\label{fig6:c}
     \end{subfigure}

\caption{Comparison of video frames generated using Dynamic Transfer GAN with different source video sequence dynamic feature encodings.}
\label{fig:6}
\vspace{-0.5cm}
\end{figure}

\begin{figure*}[!ht]
\begin{center}
\includegraphics[width=1\linewidth] {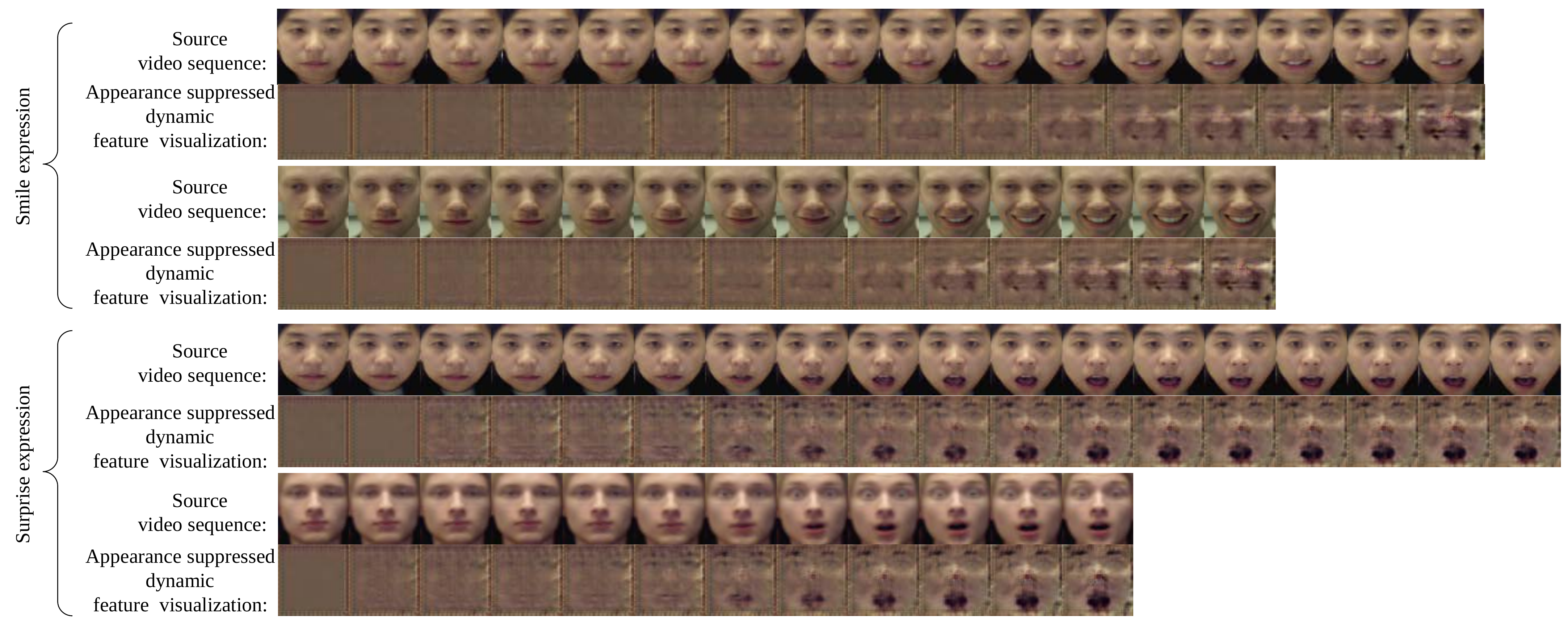}
\end{center}
   \caption{Visualization of the appearance suppressed dynamic feature.}
   \vspace{-0.5cm}
\label{fig:7}
\end{figure*}

\subsection{Experiment 3: Comparison with Different Source Video Dynamic Feature Encodings}
In this experiment, we investigated the quality of the generated video sequences with the Dynamics Transfer GAN by using different dynamic feature encoding. To that end, we generated video sequences using the proposed Dynamics Transfer GAN that utilizes the proposed appearance suppressed dynamics feature  $(\textbf{\textup{Z}}^{(d)}=\textbf{\textup{H}}-\textbf{\textup{H}}^{(st)})$as detailed in Figure~\ref{fig:3} . For comparison, different dynamic feature encoding was performed by replacing the appearance suppressed dynamics feature with the CNN-LSTM features \cite{24} (the source video spatio-temporal feature in Figure~\ref{fig:3} is used as input of the dynamic channel embedder $(F)$ in Figure~\ref{fig:4} , i.e., $\textbf{\textup{Z}}^{(d)}=\textbf{\textup{H}}$ ). 

Figure~\ref{fig:6} shows a number of example frames from the generated video sequences using the proposed Dynamics Transfer GAN. Figure~\ref{fig6:a}  shows frames from the source video sequence. The images in Figure~\ref{fig6:b} were generated using the proposed Dynamic Transfer GAN with proposed appearance suppressed dynamics features $(\textbf{\textup{Z}}^{(d)}=\textbf{\textup{H}}-\textbf{\textup{H}}^{(st)})$. The images in Figure~\ref{fig6:c}  were generated using the proposed Dynamic Transfer GAN with the CNN-LSTM features \cite{24}. For more examples, please refer to the appendix section 6.2. By inspecting the location of the artifacts in the images generated using the CNN-LSTM features, we observed that artifacts could mainly occur at: (1) frames with larger dynamics (frames with intense expressions) and (2) locations where there was a deformation in the spatial construct of the source video frame (e.g., wrinkle locations). On the other hand, such artifacts were minimized when the Dynamics Transfer GAN utilized the appearance suppressed dynamics features. These results show that the proposed appearance suppressed dynamics feature could encode the dynamics of the source features more efficiently. As a result, the generated images have fewer artifacts compared to the model utilizing the CNN-LSTM features.

We further performed a subjective preference experiment, to quantitatively evaluate the effect of the source video sequence dynamic encoding method on the generated videos. To that end, 10 participants were summoned. This time, the subjects were presented with 240 pairs of video sequences. In each pair, one video was generated using the proposed appearance suppressed dynamics feature. The other video was generated using the CNN-LSTM features. The subjects were asked to state which video sequence they preferred, in terms of quality. Note that to avoid bias in the subject's decisions, the location of the presented pair of sequences was switched randomly. Table~\ref{Table:2} shows that the videos generated using the appearance suppressed dynamics feature were overwhelmingly preferred by the test subjects. These results support the fact that the appearance suppressed dynamics feature is more efficient in encoding the dynamics of the source video sequence compared to the spatio-temporal features of the CNN-LSTM \cite{24}. It also confirms the qualitative results that generated video sequence generated via the appearance suppressed dynamics feature are less prone to artifacts.

\begin{table}[!ht]
  \centering
  \caption{Subject preference results for the  generated videos (the proposed appearance suppressed dynamics features vs. CNN-LSTM features)}
  \label{Table:2}
  \resizebox{\columnwidth}{!}{
    \begin{tabular}{p{8.33cm}rrr}
    \toprule
    \toprule
    \multicolumn{1}{p{1.33cm}}{} & \multicolumn{1}{p{2.cm}}{\textbf{CNN-LSTM \newline{} features }} & \multicolumn{1}{p{3.8cm}}{\textbf{Appearance suppressed \newline{} dynamics features}} \\
   
    \midrule
    \multicolumn{1}{p{2.2cm}}{Preference (\%)} & \multicolumn{1}{c}{17.12\%} & \multicolumn{1}{c}{82.88\%} \\
    \midrule
    \midrule

    \end{tabular}%
    }
\end{table}%
\vspace{-0.3cm}

\subsection{Experiment 4: Visualization of the Appearance Suppressed Dynamics Features}
In experiment 4, we intend to visualize the appearance suppressed dynamics feature of different source video sequence. After the Dynamics Transfer GAN is trained, a video sequence is generated by importing the dynamics of a source video sequence, and imposing them on a target image. However, we wondered what would happen if the target image did not contain any spatial construct (i.e., an image with zero pixel values). This should provide a visualization for the appearance suppressed dynamics features of the source video. To test that hypothesis, video sequences were constructed with a target image of no spatial construct (image with zero pixel values). Examples of the resulting video sequences are shown in Figure~\ref{fig:7}. As shown in the figure, the generated video sequences share the same dynamics with the source video sequence. However, the appearance (identity of the subject) in the source sequence was removed. Another way to interpret these results is that the generator constructed a visualization of the appearance suppressed dynamics features (which was obtained from the source video sequence via the RNN). This visualization validates that the proposed appearance suppressed dynamics features suppress the spatial appearance of the source video sequence.

\section{Conclusion}
In this paper, we proposed a new video generation method based on GAN. The proposed method transfers arbitrary dynamics from a source video sequence to a target image. The spatial construct of a generated video sequence acquires the appearance of the target image. To achieve that, the appearance of the source video sequence was suppressed and only the dynamics of the source video were obtained before being imposed onto the target image. Therefore, an appearance suppressed dynamics feature was proposed to diminish the appearance of source video sequence while strictly encoding the dynamics. The appearance suppressed dynamics features utilized a pre-trained RNN network. Two discriminators were proposed: the spatial discriminator to validate the fidelity of the generated frames appearance and the dynamics discriminator to validate the continuity of the generated video dynamics. 

Qualitative and subjective experiments have been conducted to verify the quality of the generated videos with the proposed Dynamics Transfer GAN. The results verified that the proposed method was able to impose the arbitrary dynamics of a source video sequence onto a target image, without distorting the spatial construct of that image. The experiments also verified that Dynamics Transfer GAN generates spatially and temporally consistent video sequences.


{\small
\bibliographystyle{ieee}
\bibliography{refs}
}
\section{Appendix}
\subsection{Additional Examples of Sequences Generated Using the Proposed Dynamics Transfer GAN}
\begin{figure*}[!ht]
\begin{center}
\includegraphics[angle=270,width=9.0cm,keepaspectratio] {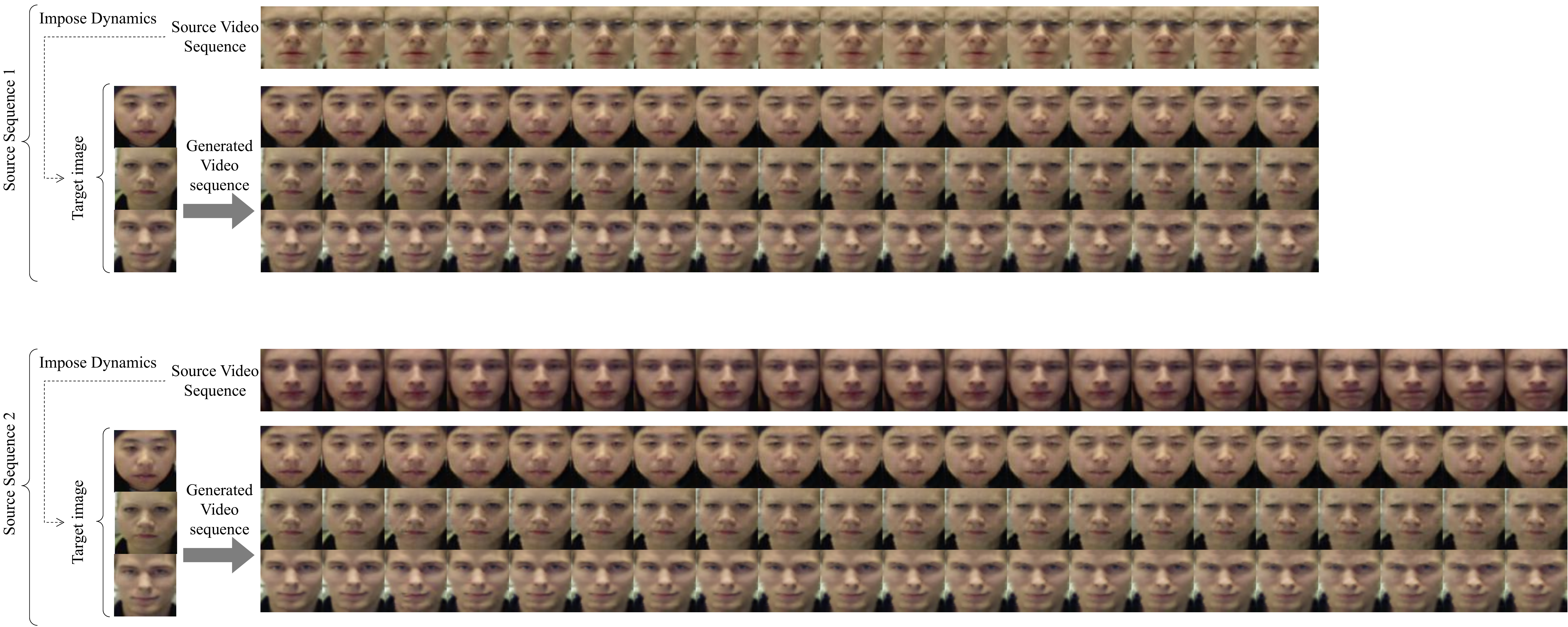}
\end{center}
   \caption{Examples of video sequences (anger expression) generated with the proposed Dynamics Transfer GAN.}
\label{Appfig:1}
\end{figure*}

\begin{figure*}[!ht]
\begin{center}
\includegraphics[angle=270,width=9.0cm,keepaspectratio] {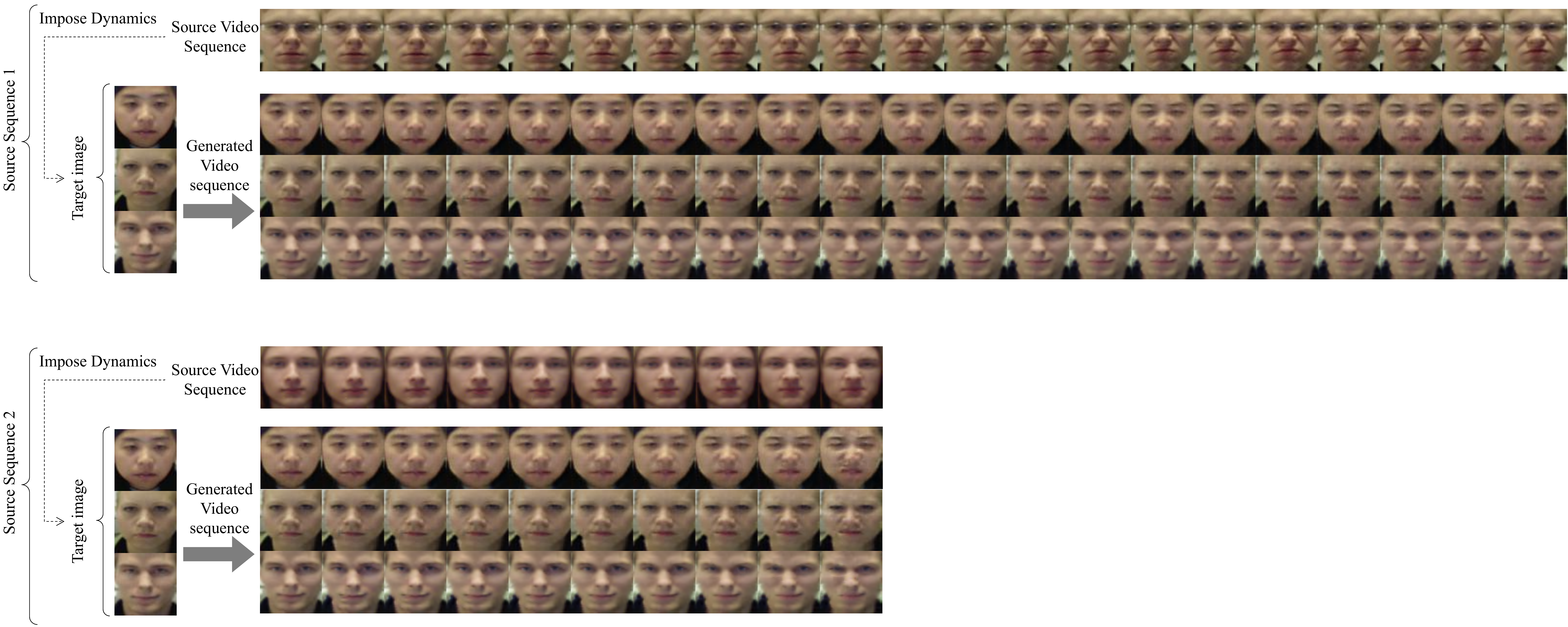}
\end{center}
   \caption{Examples of video sequences (disgust expression) generated with the proposed Dynamics Transfer GAN.}
\label{Appfig:2}
\end{figure*}

\begin{figure*}[!ht]
\begin{center}
\includegraphics[angle=270,width=9.0cm,keepaspectratio] {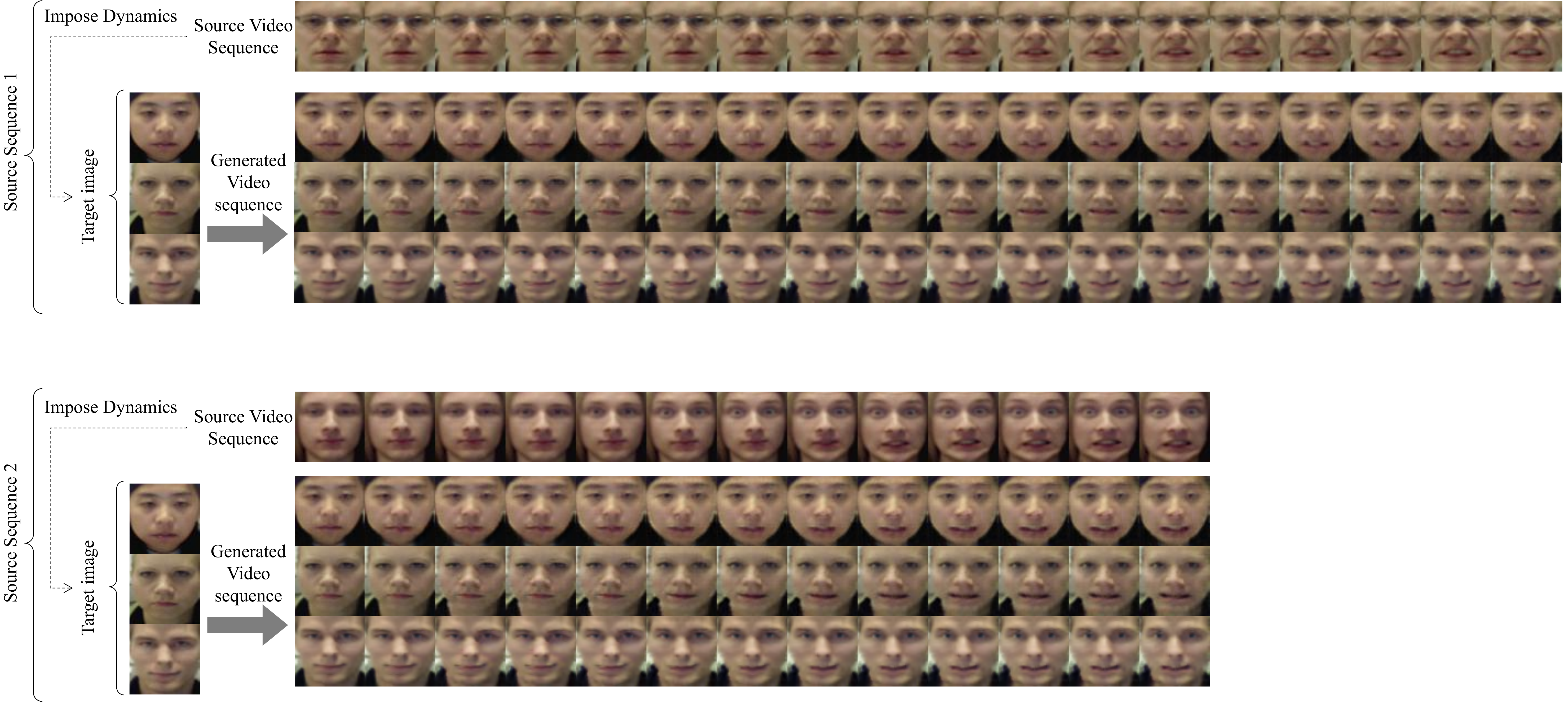}
\end{center}
   \caption{Examples of video sequences (fear expression) generated with the proposed Dynamics Transfer GAN.}
\label{Appfig:3}
\end{figure*}

\begin{figure*}[!ht]
\begin{center}
\includegraphics[angle=270,width=8.0cm,keepaspectratio] {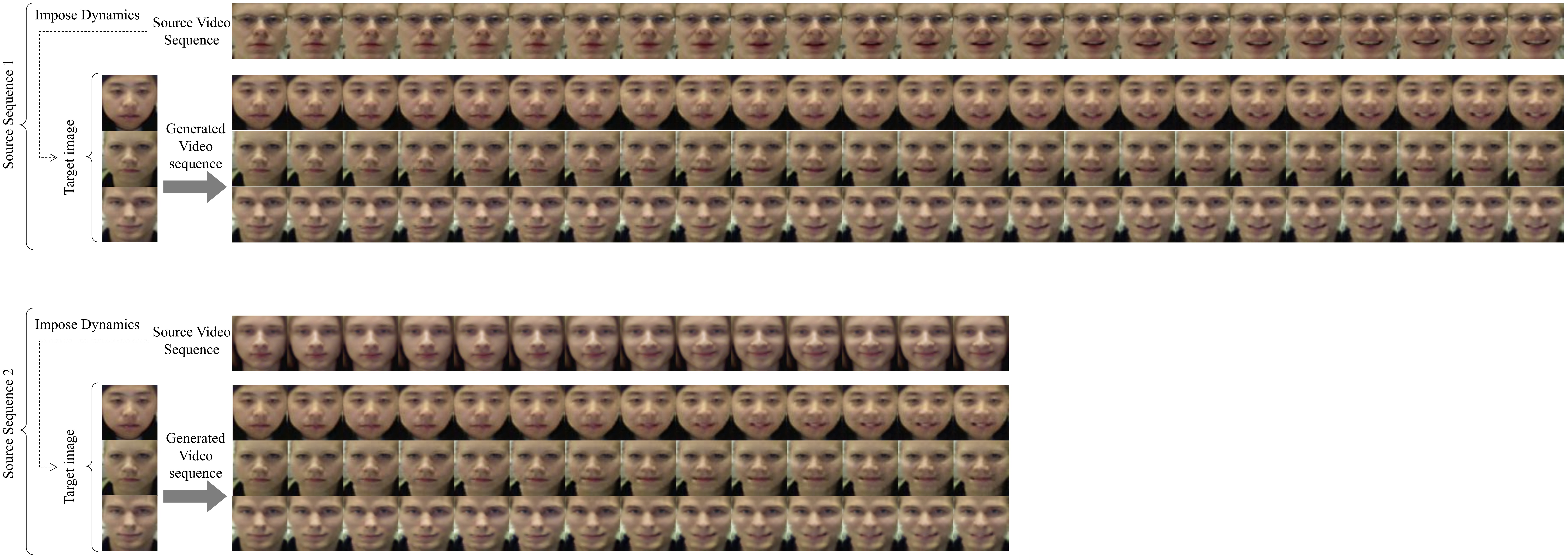}
\end{center}
   \caption{Examples of video sequences (happiness expression) generated with the proposed Dynamics Transfer GAN.}
\label{Appfig:4}
\end{figure*}

\begin{figure*}[!ht]
\begin{center}
\includegraphics[angle=270,width=9.0cm,keepaspectratio] {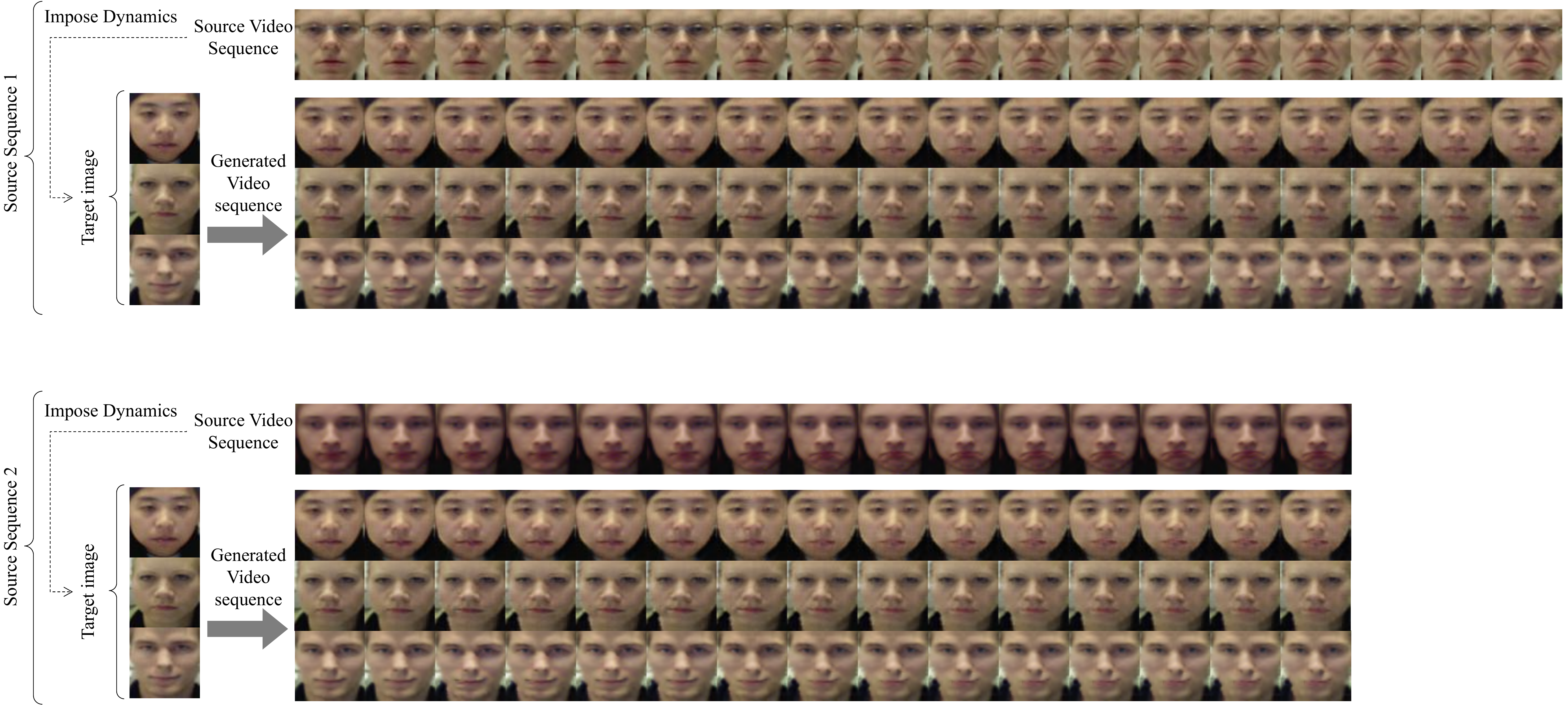}
\end{center}
   \caption{Examples of video sequences (sadness expression) generated with the proposed Dynamics Transfer GAN.}
\label{Appfig:5}
\end{figure*}

\begin{figure*}[!ht]
\begin{center}
\includegraphics[angle=270,width=8.0cm,keepaspectratio] {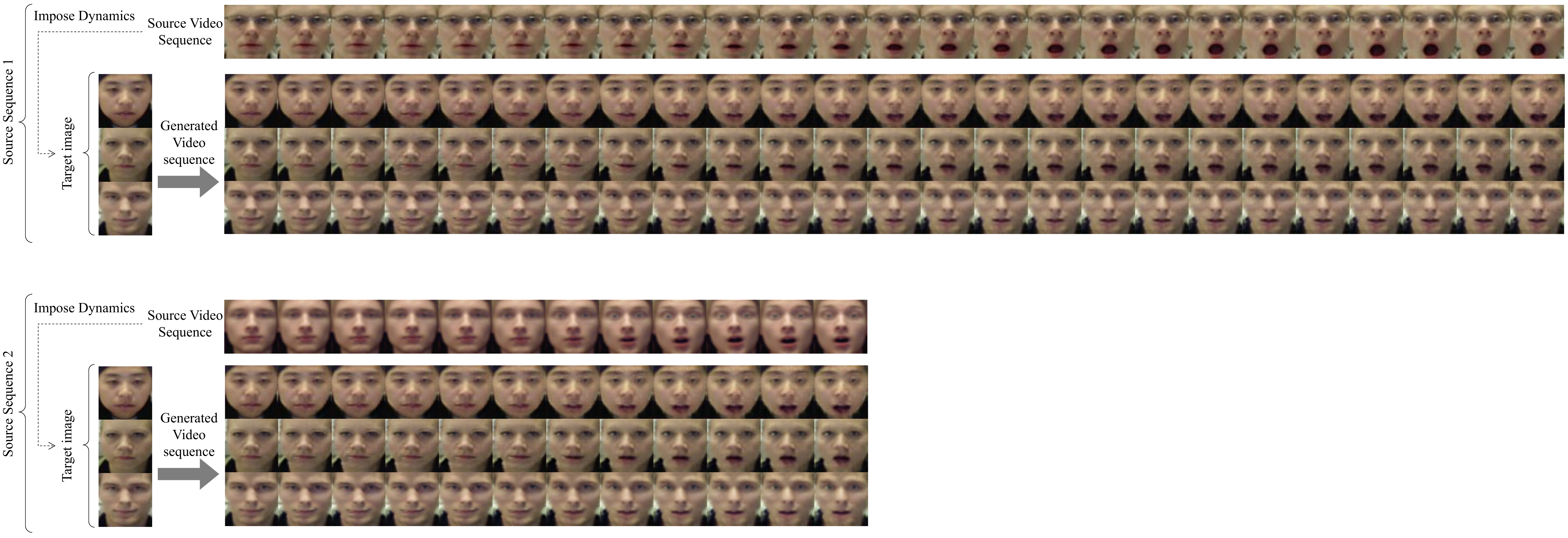}
\end{center}
   \caption{Examples of video sequences (surprise expression) generated with the proposed Dynamics Transfer GAN.}
\label{Appfig:6}
\end{figure*}

In the main paper (particularly section 3.2), we showed a few examples of generated video sequences using the proposed Dynamics Transfer GAN. In Figures \ref{Appfig:1},\ref{Appfig:2},\ref{Appfig:3},\ref{Appfig:4},\ref{Appfig:5} and \ref{Appfig:6}, we show additional examples of different generated facial expression sequences. In each figure, two different source video sequences, three target images and the corresponding generated video sequences. Please refer to the video \url{https://youtu.be/ppAUF1WVur8}to watch playable examples of generated video sequences.

\subsection{Additional Comparison Examples with Different Source Video Dynamic Feature Encodings}
\begin{figure*}[!ht]
    \centering
    
    \begin{subfigure}[b]{0.45\textwidth}
                 \centering
                 \includegraphics[width=1\textwidth,keepaspectratio]{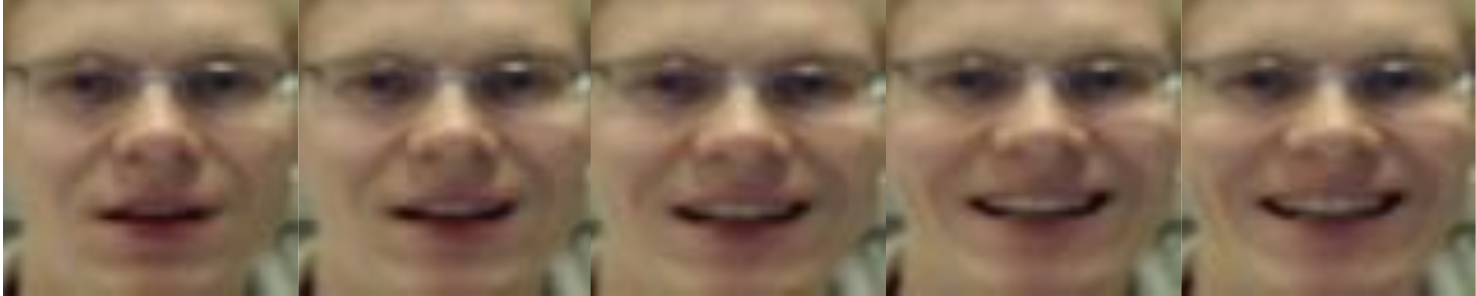}
                 \caption{Samples from a smile expression source video sequence.}
             	 \label{Appfig7:a}
         \end{subfigure}
         \begin{subfigure}[b]{0.45\textwidth}
                 \centering
                 \includegraphics[width=1\textwidth,keepaspectratio]{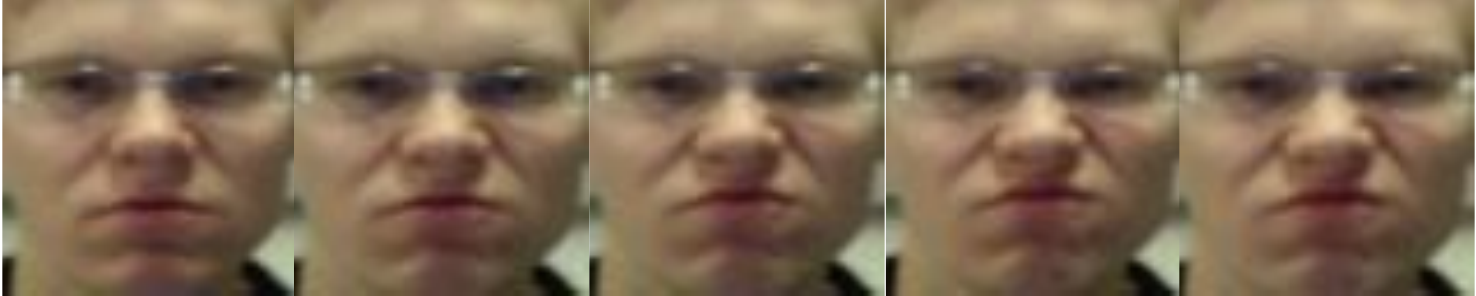}
                 \caption{Samples from a disgust expression source video sequence.}
             	 \label{Appfig7:b}
         \end{subfigure}%

         \begin{subfigure}[b]{0.45\textwidth}
                 \centering
                 \includegraphics[width=1\textwidth,keepaspectratio]{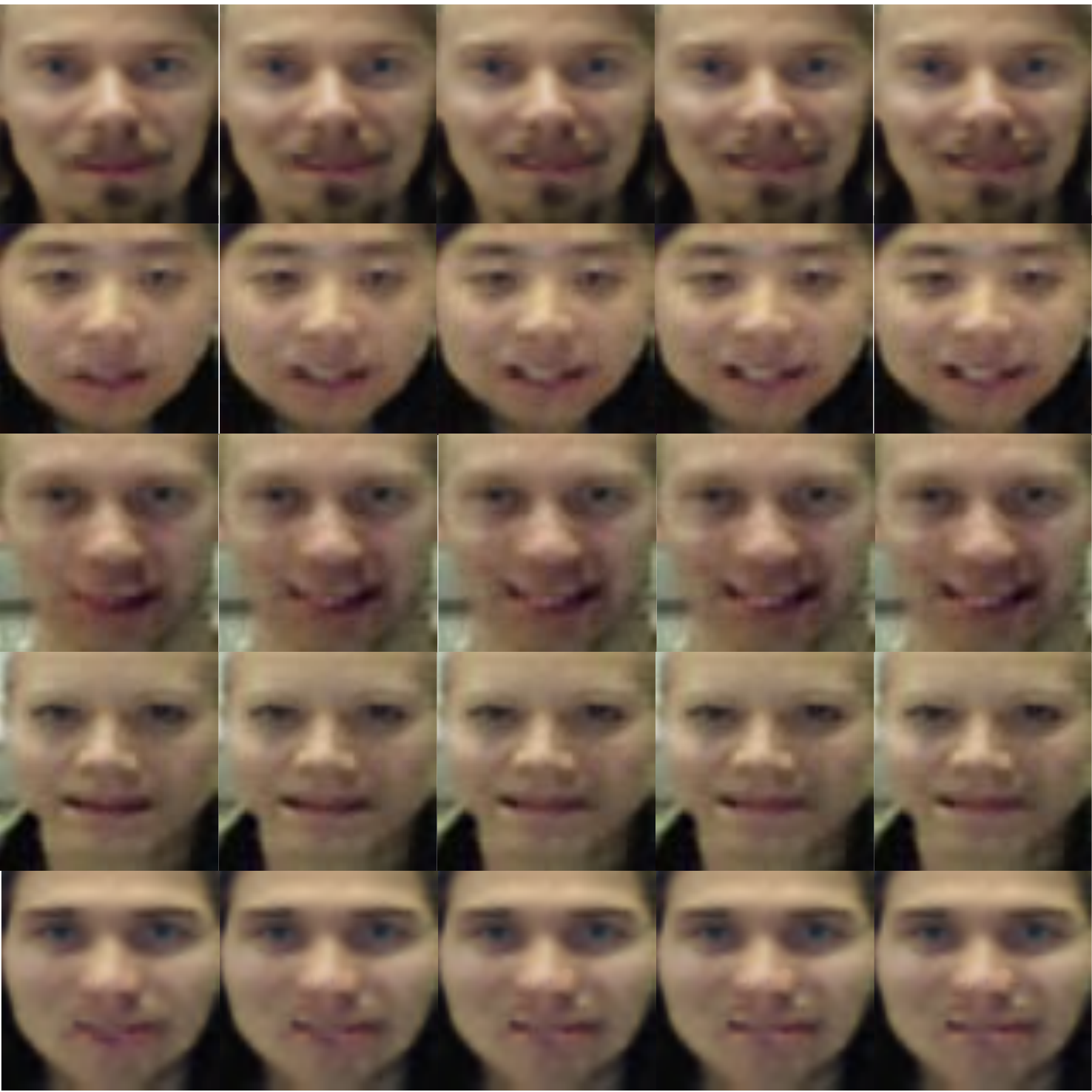}
                 \caption{Samples generated with Dynamics Transfer GAN using the proposed appearance suppressed Dynamics feature. Note that the source sequence is shown in (a).}
             	 \label{Appfig7:c}
         \end{subfigure}
     \begin{subfigure}[b]{0.45\textwidth}
             \centering
             \includegraphics[width=1\textwidth,keepaspectratio]{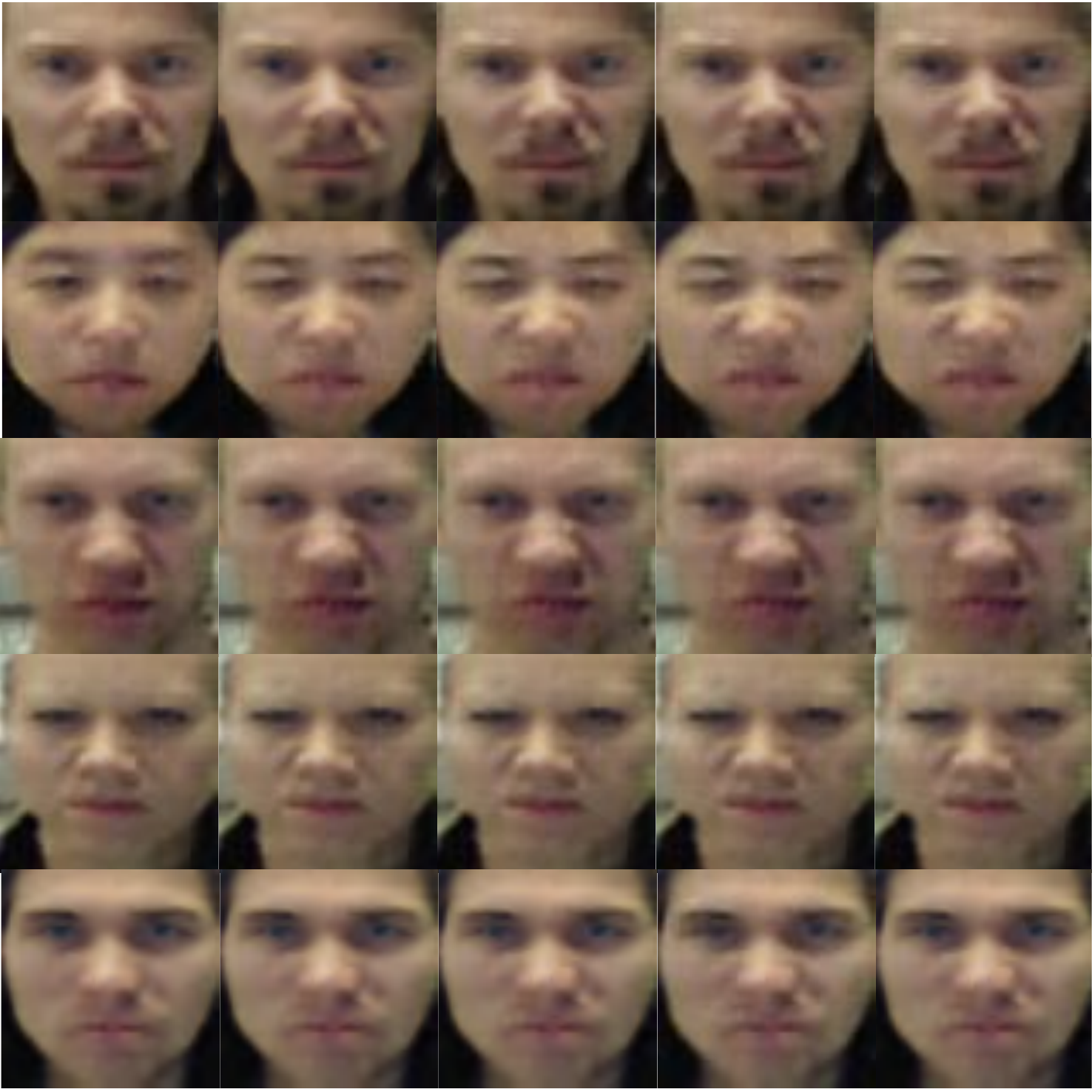}
             \caption{Samples generated with Dynamics Transfer GAN using the proposed appearance suppressed Dynamics feature. Note that the source sequence is shown in (b).}
             \label{Appfig7:d}
     \end{subfigure}

     \begin{subfigure}[b]{0.45\textwidth}
             \centering
             \includegraphics[width=1\textwidth,keepaspectratio]{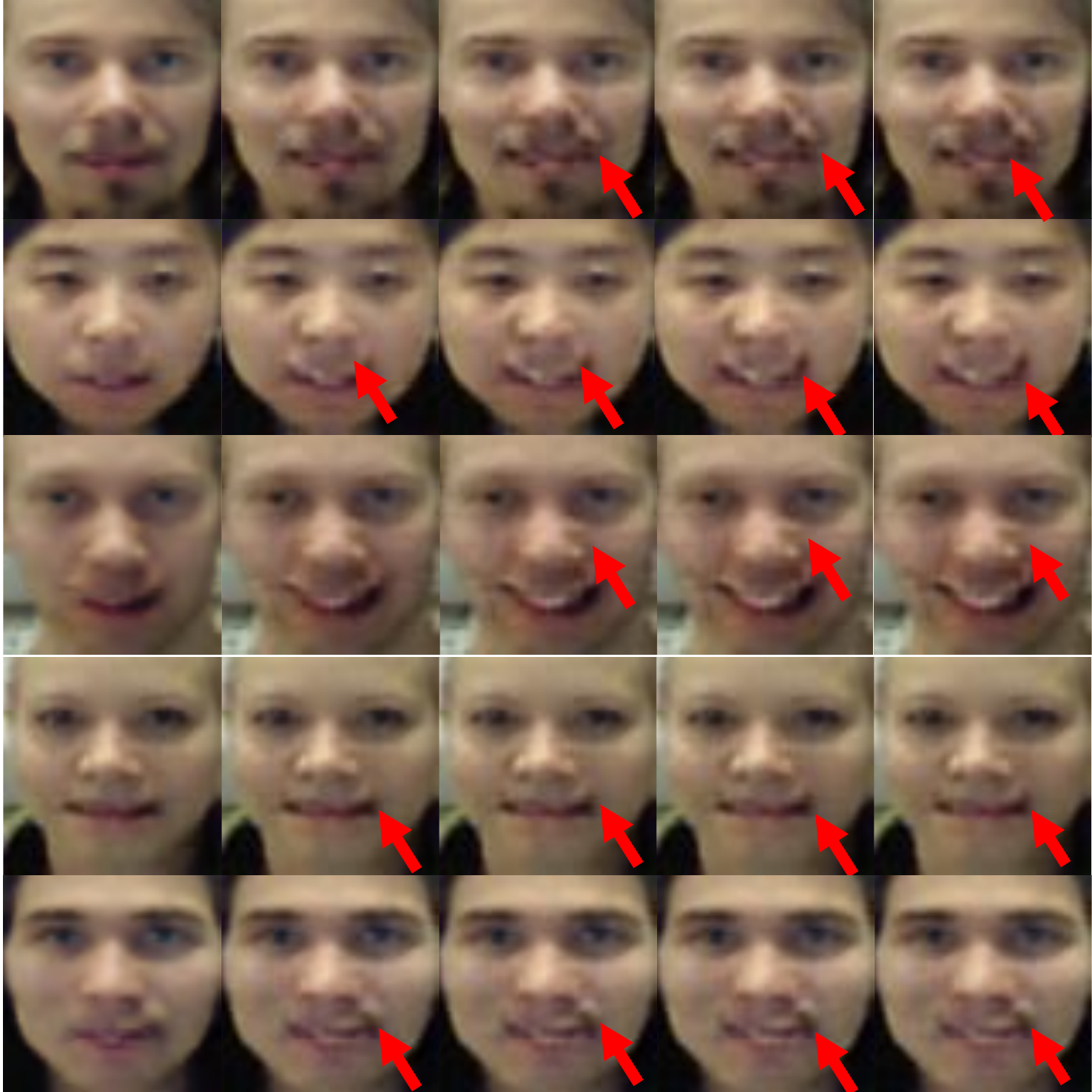}
             \caption{Samples generated using CNN-LSTM features. Note that the source sequence is shown in (a).}
             \label{Appfig7:e}
      \end{subfigure}
      \begin{subfigure}[b]{0.45\textwidth}
             \centering
             \includegraphics[width=1\textwidth,keepaspectratio]{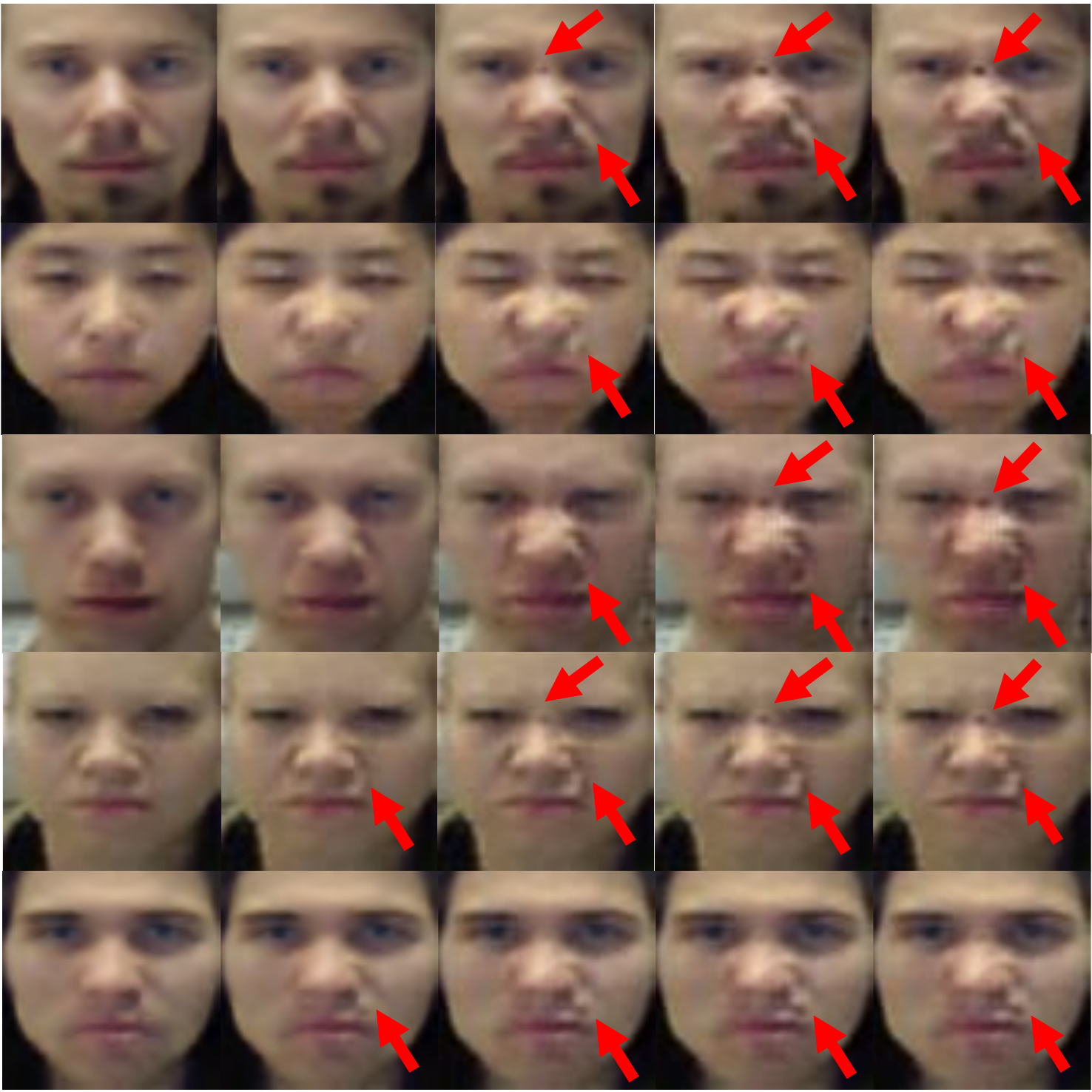}
             \caption{Samples generated using CNN-LSTM features. Note that the source sequence is shown in (b).}
             \label{Appfig7:f}
      \end{subfigure}

\caption{Comparison of video frames generated using Dynamic Transfer GAN with different source video sequence dynamic feature encodings.}
\label{Appfig:7}
\end{figure*}

In the section 3.4 of the main paper, we showed some comparative examples of video frames generated using Dynamic Transfer GAN with different source video sequence dynamic feature encodings. In Figures~\ref{Appfig:7}, additional examples are shown with different subjects and facial expressions. Please refer to the video at \url{https://youtu.be/ppAUF1WVur8} to watch playable examples of comparison example video sequences.

\subsection{Examples of Generated Video Sequences Imported from Long Source Video Sequences}
In the video \url{https://youtu.be/ppAUF1WVur8}, we provide examples of generated lengthy videos sequences (large number of frames). The dynamics of a generated video was imported from one long video sequence composed of the 6 basic expressions performed continuously. From the results shown in the video examples, the proposed Dynamics Transfer GAN generated video sequences with reasonable quality regardless of the lengthy sequences.

\end{document}